\documentclass[conference]{IEEEtran}
\IEEEoverridecommandlockouts
\usepackage{cite}
\usepackage{amsmath,amssymb,amsfonts}
\usepackage{algorithmic}
\usepackage{graphicx}
\usepackage{textcomp}
\usepackage{xcolor}
\usepackage{booktabs}
\usepackage{multirow}
\usepackage{subcaption}
\usepackage{pgfplots}
\usepackage{float}
\usepackage{dblfloatfix}
\pgfplotsset{compat=1.17}
\usepackage[font=small,labelfont=bf]{caption}

\def\BibTeX{{\rm B\kern-.05em{\sc i\kern-.025em b}\kern-.08em
    T\kern-.1667em\lower.7ex\hbox{E}\kern-.125emX}}
\begin{document}

\title{Accelerating Bangla NLP Tasks with Automatic Mixed Precision: Resource-Efficient Training Preserving Model Efficacy}

\author{
\IEEEauthorblockN{Md Mehrab Hossain Opi}
\IEEEauthorblockA{
\textit{Department of Computer Science and Engineering} \\
\textit{Khulna University of Engineering \& Technology} \\
Khulna, Bangladesh \\
Email: opi@cse.kuet.ac.bd
}
\and
\IEEEauthorblockN{Sumaiya Khan}
\IEEEauthorblockA{
\textit{Department of Computer Science and Engineering} \\
\textit{Khulna University of Engineering \& Technology} \\
Khulna, Bangladesh \\
Email: sumaiyakhan3572213@gmail.com
}
\and
\IEEEauthorblockN{Moshammad Farzana Rahman}
\IEEEauthorblockA{
\textit{Department of Computer Science and Engineering} \\
\textit{Khulna University of Engineering \& Technology} \\
Khulna, Bangladesh \\
Email: rfarzana964@gmail.com
}
}

\maketitle

\begin{abstract}
Training models for Natural Language Processing (NLP) requires substantial computational resources and time, posing significant challenges, especially for NLP development in Bangla, where access to high-end hardware is often limited. In this work, we explore automatic mixed precision (AMP) training as a means to improve computational efficiency without sacrificing model performance. By leveraging a dynamic mix of 16-bit and 32-bit floating-point computations, AMP lowers GPU memory requirements and speeds up training without degrading model performance. We evaluate AMP across four standard Bangla NLP tasks, namely sentiment analysis, named entity recognition, error classification, and question answering, using four transformer-based models: BanglaBERT, BanglishBERT, XLM-R, and mBERT. Our results demonstrate that AMP accelerates training by 44.5\% and reduces memory consumption by 17.6\%, while maintaining F-1 score within 99.7\% of the full-precision baselines. This empirical study highlights AMP's potential to democratize access to state-of-the-art NLP capabilities in hardware-constrained settings by lowering computational barriers.
\end{abstract}

\begin{IEEEkeywords}
HPC, NLP, AMP, Benchmarking
\end{IEEEkeywords}
\section {Introduction}
Natural Language Processing (NLP) has become a foundational subfield of artificial intelligence, enabling applications such as machine translation, sentiment analysis, and conversational agents, and has emerged as a core research area powering systems from search engines to intelligent virtual assistants\cite{Khurana2017NaturalLP}. 
Recent breakthroughs in Large Language Models (LLMs) have catalyzed further progress, demonstrating unprecedented task performance across multiple NLP domains\cite{Kumar2024LargeLM}. 
Despite notable progress, research has largely centered on high-resource languages, leaving Bangla, one of the world’s most widely spoken languages, comparatively underrepresented. 
Emerging pretrained Bangla language models\cite{Bhattacharjee2021BanglaBERTLM, Bhattacharjee2022BanglaNLGAB, Bhattacharyya2025BanglaByT5BM, Salim2023BanglaGPTAG} have demonstrated strong potential on downstream tasks. However, the training and deployment of large-scale NLP models are computationally demanding, necessitating substantial memory and processing capacity on modern GPUs. These requirements pose significant challenges for researchers and organizations working with limited hardware resources, a situation common in developing regions where Bangla NLP research is gaining momentum, emphasizing the urgent need for efficiency-oriented solutions.

Mixed-precision training\cite{Micikevicius2017MixedPT} has emerged as an effective approach to address these computational challenges. It leverages a combination of 16-bit floating-point (FP16) and 32-bit floating-point (FP32) data types during training. By performing most computations in the smaller FP16 format, mixed-precision training effectively halves memory usage and can significantly accelerate matrix multiplication operations, which are the backbone of modern deep learning. This enables models to leverage lower-precision arithmetic without compromising accuracy. While AMP has demonstrated substantial improvements in training speed and memory efficiency for high-resource language models, to the best of our knowledge, no systematic study has yet investigated its impact on Bangla NLP tasks. This leaves a critical gap in understanding its effectiveness in this important yet under-explored context, especially for overcoming the hardware limitations faced by local researchers.

In this work, we investigate the effectiveness of Automatic Mixed Precision for Bangla NLP tasks using GPU-based training. We select a few widely used pretrained Bangla language models as the baseline and apply AMP to accelerate training while reducing memory overhead. The objective is to evaluate whether AMP can deliver measurable improvements in training efficiency, specifically in training time and GPU memory utilization without degrading task performance. To achieve this, we conduct experiments on four standard Bangla NLP tasks using widely-used Transformer-based models. We systematically compare full-precision and AMP-enabled training across multiple performance metrics. Our findings demonstrate that AMP significantly reduces training time and memory consumption while maintaining model accuracy.

The main contributions of this work are as follows:
\begin{itemize}
    \item We implement automatic mixed precision (AMP) training on multiple pretrained Transformer models for Bangla NLP tasks.
    \item We systematically evaluate AMP across four representative Bangla NLP tasks to quantify efficiency and performance trade-offs.
    \item We demonstrate substantial efficiency gains with AMP, including reduced GPU memory usage and faster training, while maintaining model performance.
    \item We provide practical guidance for optimizing training workflows in low-resource Bangla NLP settings without compromising accuracy.
\end{itemize}

\section{Related Work}
Research in Bangla NLP has advanced considerably with the development of pretrained language models tailored to the language. BERT-based architectures, such as BanglaBERT and BanglishBERT \cite{Bhattacharjee2021BanglaBERTLM}, have been successfully applied to various NLP tasks, trained on large Bangla corpora. Recent encoder-decoder models, including BanglaT5 \cite{Bhattacharjee2022BanglaNLGAB} and BanglaByT5 \cite{Bhattacharyya2025BanglaByT5BM}, facilitate low-resource natural language generation in Bangla, while BanglaGPT \cite{Salim2023BanglaGPTAG} demonstrates generative pretrained transformer capabilities. Multilingual transformers such as mBERT \cite{Devlin2019BERTPO}, MuRIL \cite{khanuja2021muril}, and XLM-RoBERTa (XLM-R) \cite{Conneau2019UnsupervisedCR} achieve competitive performance for Bangla despite being trained on multiple languages. More recently, Bangla-focused large language models (LLMs), including BongLLaMA \cite{Zehady2024BongLLaMALF} and TigerLLM \cite{Raihan2025TigerLLMA}, indicate growing efforts to bring LLM capabilities to Bangla NLP. Despite their strong performance, these models require substantial GPU resources, limiting accessibility in low-resource settings.

Mixed-precision training \cite{Micikevicius2017MixedPT} has emerged as a widely adopted technique for accelerating deep learning by combining lower-precision arithmetic (e.g., FP16) with standard precision (FP32). Automatic Mixed Precision (AMP), available in frameworks such as PyTorch and TensorFlow, improves training speed and reduces memory consumption without degrading model accuracy in high-resource language tasks \cite{Kuchaiev2018MixedPrecisionTF, Zhao2021AutomaticMQ}. While AMP has been applied to large-scale NLP models in English and other widely studied languages, its potential in Bangla NLP remains unexplored.

In summary, although pretrained Bangla models exist and AMP has proven effective in other languages, systematic studies applying AMP to Bangla NLP tasks are missing. Our work addresses this gap by evaluating AMP across multiple Bangla NLP benchmarks, measuring both task performance and training efficiency, thereby providing practical insights for researchers working with limited computational resources.

\section{Methodology}

This study investigates the efficiency of Automatic Mixed Precision (AMP) training across multiple Bangla NLP tasks using pretrained transformer architectures. We design a unified training and evaluation pipeline to enable consistent comparison between full precision (FP32) and mixed precision (FP16/FP32) training. Four representative tasks are selected to cover sentence-level, token-level, and span-level NLP problems. Pretrained monolingual and multilingual transformer models—including BanglaBERT\cite{Bhattacharjee2021BanglaBERTLM}, BanglishBERT\cite{Bhattacharjee2021BanglaBERTLM}, mBERT\cite{Devlin2019BERTPO}, and XLM-R\cite{Conneau2019UnsupervisedCR}—are fine-tuned under both training regimes. Performance is evaluated using accuracy- and efficiency-oriented metrics, highlighting the balance between computational resource usage and task performance.

\subsection{Task Description and Dataset Overview}

We evaluate AMP training on four Bangla NLP tasks:

\begin{itemize}
  \item \textbf{Sentiment Analysis (SA):} Classifies text into positive, negative, or neutral sentiment using the BLP-2023\cite{Hasan2023BLP2023T2} dataset, which integrates \textbf{MUBASE}\cite{Hasan2023ZeroAF} and \textbf{SentiNob}\cite{Islam2021SentNoBAD}, comprising 46,643 examples. Preprocessing includes removal of URLs, emojis, and special characters, followed by model-specific tokenization using SentencePiece.
    \item \textbf{Named Entity Recognition (NER):} Identifies named entities in Bangla health-related text using the \textbf{Bangla-HealthNER} dataset\cite{Khan2023NERvousAM}, containing 31,783 annotated samples with seven entity types in IOB format. The dataset includes code-switched Bangla-English text.
    \item \textbf{Error Classification (EC):} Detects and categorizes errors in user-generated Bangla text using the \textbf{BaTEClaCor}\cite{Oshin2023BaTEClaCorAN} dataset, comprising 10,000 YouTube comments across multiple domains. Only comments with at least three Bangla words are included.
    \item \textbf{Question Answering (QA):} Answers questions based on Bangla passages using the \textbf{BanglaRQA} dataset\cite{Ekram2022BanglaRQAAB}, containing 14,889 question–answer pairs over 3,000 Wikipedia passages. Questions are of four types (list, factoid, causal, confirmation) and answers of three types (yes/no, single span, multiple spans).
\end{itemize}
Preprocessing for all tasks includes normalization, tokenization, and truncation/padding to fit model input requirements.

\subsection{Model Selection}

Four pretrained transformer models are selected to represent monolingual and multilingual approaches:

\begin{itemize}
    \item \textbf{BanglaBERT:} BERT-based model pretrained on Bangla text \cite{Bhattacharjee2021BanglaBERTLM}.
    \item \textbf{BanglishBERT:} Trained on code-mixed Bangla-English text to capture social media linguistic nuances \cite{Bhattacharjee2021BanglaBERTLM}.
    \item \textbf{mBERT:} Multilingual BERT pretrained on 104 languages including Bangla, offering cross-lingual capabilities \cite{Devlin2019BERTPO}.
    \item \textbf{XLM-R:} Multilingual transformer pretrained on 100+ languages, known for robust cross-lingual transfer \cite{Conneau2019UnsupervisedCR}.
\end{itemize}

\subsection{Methodology Pipeline}

\begin{figure}[htbp]
    \centering
    \includegraphics[width=0.7\linewidth]{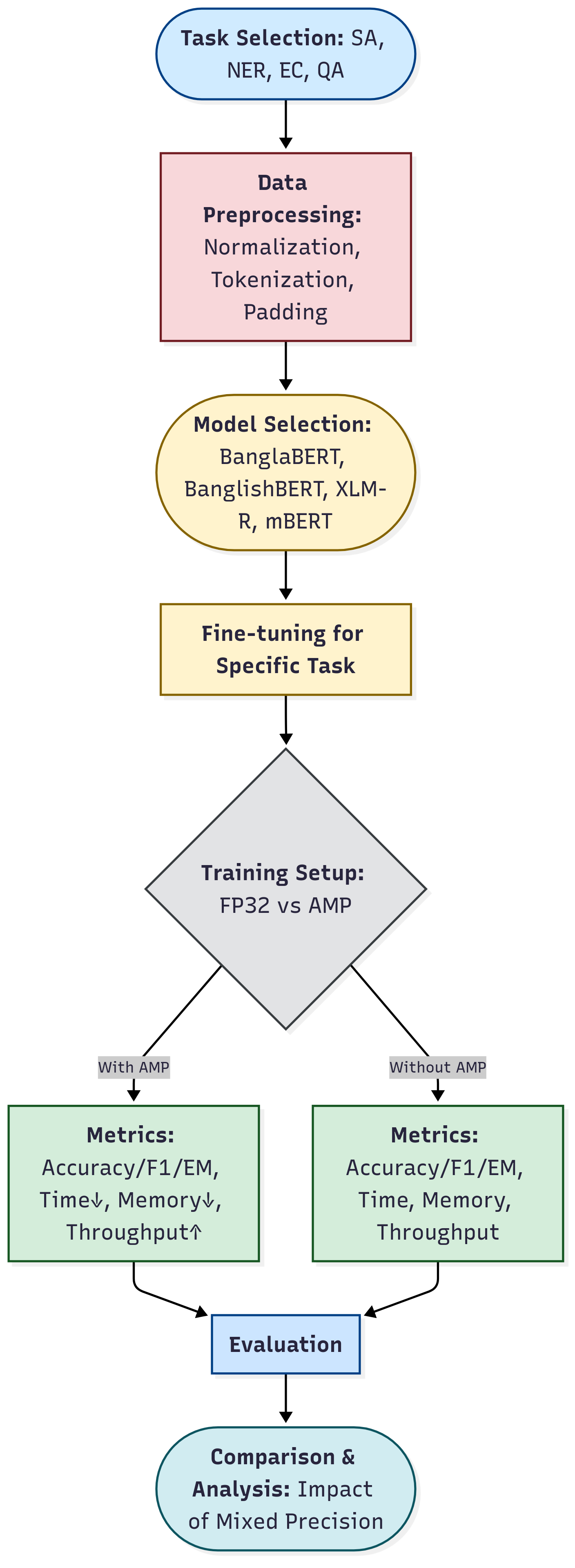}
    \caption{Overview of the proposed AMP training methodology for Bangla NLP tasks.}
    \label{fig:methodology_flowchart}
\end{figure}

The training pipeline, illustrated in Figure~\ref{fig:methodology_flowchart}, follows these steps:

\begin{enumerate}
    \item \textbf{Task Selection:} Four representative Bangla NLP tasks are selected to evaluate AMP training.
    \item \textbf{Data Preprocessing:} Text is normalized using the \texttt{normalizer} library, tokenized with model-specific tokenizers, and padded to maximum sequence lengths with attention masks.
    \item \textbf{Model Initialization:} Pretrained transformers are augmented with task-specific classification or span prediction heads; token-level classification is used for NER and EC tasks.
    \item \textbf{Training Setup:} Models are trained under FP32 and AMP using AdamW with task-specific learning rates and weight decay. Gradient clipping and early stopping are applied based on validation performance. Batch sizes and maximum sequence lengths are given in Table~\ref{tab:training_config_full}.
    \item \textbf{Evaluation:} Task performance is measured using Accuracy, F1-score, and Exact Match (EM), while efficiency is assessed via peak GPU memory, epoch time, and throughput.
    \item \textbf{Comparison and Analysis:} FP32 and AMP results are systematically compared to quantify trade-offs between model performance and computational efficiency.
\end{enumerate}

\subsection{Implementation Details}

Experiments are implemented in PyTorch with the Hugging Face Transformers library. Token sequences are padded to model-specific maximum lengths with attention masks applied to distinguish real tokens from padding. Gradient clipping stabilizes training, and model checkpoints are saved at regular intervals. AMP training is facilitated via PyTorch’s \texttt{torch.cuda.amp} module, reducing memory usage and accelerating training while maintaining accuracy comparable to FP32.  

All experiments are conducted on an Intel Core i5-13500 CPU with an NVIDIA RTX 4070 GPU and 64 GB DDR4 RAM running Ubuntu 22.04. Task-specific batch sizes, learning rates, and epochs are summarized in Table~\ref{tab:training_config_full}.

\begin{table*}[htbp]

\centering

\caption{Task-Specific Model and Training Configurations}

\label{tab:training_config_full}

\footnotesize

\begin{tabular}{l l c c c c c c}

\toprule

\textbf{Task} & \textbf{Optimizer} & \textbf{Weight Decay} & \textbf{Learning Rate} & \textbf{Batch Size} & \textbf{Epochs} & \textbf{Max Seq Len} & \textbf{Dropout} \\

\midrule

Sentiment Analysis & AdamW & 0.01 & 5e-5 & 16 & 3  & 512 & 0.1 \\

NER                & AdamW & 0.03 & 1e-4 & 64 & 20 & 512 & 0.4 \\

Error Classification & AdamW & 0.01 & 1e-5 & 16 & 5 & 128 & 0.1 \\

Question Answering & AdamW & 0.00 & 2e-5 & 8  & 15 & 512 & 0.1 \\

\bottomrule

\end{tabular}

\end{table*}

\section{Experiment and Result}
Our analysis investigates the impact of Automatic Mixed Precision (AMP) training on Bangla NLP tasks using pretrained transformer models. We evaluate AMP across four representative Bangla NLP tasks: Sentiment Analysis, Named Entity Recognition (NER), Error Classification, and Question Answering, using pretrained transformer models including BanglaBERT, BanglishBERT, mBERT, and XLM-R.
We focus on two main aspects: (i) \textit{task performance}, measured in terms of accuracy and F1-score for tasks of classification and sequence labeling, and Exact Match (EM) for Question Answering, and (ii) \textit{training efficiency}, evaluated through GPU memory consumption, training time, and throughput. We also examine the effect of varying batch sizes on performance. 
All experiments compare the baseline FP32 training setup against AMP-enabled training to highlight both the benefits and potential trade-offs. 

\subsection{Task Performance}

We evaluate the impact of Automatic Mixed Precision (AMP) training utilizing FP16/FP32 mixed-precision across four Bangla NLP tasks: Sentiment Analysis (SA), Named Entity Recognition (NER), Error Classification (EC), and Question Answering (QA). Pretrained transformer models, namely BanglaBERT, BanglishBERT, mBERT, and XLM-R, are fine-tuned under standard FP32 and AMP (FP16/FP32) regimes. Task performance is measured using Accuracy and F1-score for Sentiment Analysis, Named Entity Recognition, and Error Classification, and F1-score and Exact Match (EM) for Question Answering. Tables~\ref{tab:sa_perf}--\ref{tab:qa_perf} summarize the results.

Across all tasks, training using Automatic Mixed Precision preserves or slightly improves model performance relative to the full-precision baseline. F1 retention is 98.41–99.68\% for Sentiment Analysis, 94.34–99.95\% for NER, 100.73–122.52\% for Error Classification, and 99.73–105.97\% for QA, with occasional improvements likely due to implicit regularization. These results confirm that AMP maintains task efficacy while enabling substantial computational efficiency. Key observations include:

\begin{itemize}

    \item \textbf{Sentiment Analysis:} AMP maintains comparable Accuracy and F1 across all models, with BanglaBERT achieving the highest scores.

    \item \textbf{NER:} F1 scores under AMP remain stable, with XLM-R and mBERT demonstrating the highest consistency.

    \item \textbf{Error Classification:} Accuracy is preserved under AMP, while F1 shows modest improvements for some models, reflecting robust token-level classification.

    \item \textbf{Question Answering:} AMP yields comparable or slightly higher F1 and EM scores, confirming no degradation in span-level performance.

\end{itemize}

\begin{table}[H]
\centering
\caption{Sentiment Analysis Performance: FP32 vs AMP}
\label{tab:sa_perf}
\footnotesize
\begin{tabular}{l c c}
\toprule
\textbf{Model} & Accuracy (FP32 / AMP) & F1 (FP32 / AMP) \\
\midrule
BanglaBERT    & 72.70 / 72.27 & 72.69 / 72.26 \\
BanglishBERT  & 71.33 / 71.10 & 71.32 / 71.09 \\
XLM-R         & 68.25 / 67.39 & 68.25 / 67.38 \\
mBERT         & 66.34 / 65.28 & 66.34 / 65.27 \\
\bottomrule
\end{tabular}
\end{table}

\begin{table}[htbp]
\centering
\caption{NER Task Performance: FP32 vs AMP}
\label{tab:ner_perf}
\footnotesize
\begin{tabular}{l c c}
\toprule
\textbf{Model} & Accuracy (FP32 / AMP) & F1 (FP32 / AMP) \\
\midrule
BanglaBERT    & 88.52 / 88.25 & 55.78 / 54.35 \\
BanglishBERT  & 88.89 / 88.74 & 57.54 / 56.71 \\
XLM-R         & 89.08 / 89.23 & 58.50 / 58.47 \\
mBERT         & 87.99 / 87.24 & 55.03 / 51.90 \\
\bottomrule
\end{tabular}
\end{table}

\begin{table}[htbp]
\centering
\caption{Error Classification Performance: FP32 vs AMP}
\label{tab:error_perf}
\footnotesize
\begin{tabular}{l c c}
\toprule
\textbf{Model} & Accuracy (FP32 / AMP) & F1 (FP32 / AMP) \\
\midrule
BanglaBERT    & 73.88 / 72.64 & 52.88 / 54.13 \\
BanglishBERT  & 72.63 / 72.19 & 52.61 / 53.00 \\
XLM-R         & 70.15 / 70.04 & 42.29 / 51.79 \\
mBERT         & 67.86 / 67.36 & 50.57 / 51.17 \\
\bottomrule
\end{tabular}
\end{table}

\begin{table}[H]
\centering
\caption{Question Answering Performance: FP32 vs AMP}
\label{tab:qa_perf}
\footnotesize
\begin{tabular}{l c c}
\toprule
\textbf{Model} & F1 (FP32 / AMP) & EM (FP32 / AMP) \\
\midrule
BanglaBERT    & 57.95 / 60.10 & 41.46 / 44.60 \\
BanglishBERT  & 55.02 / 56.28 & 37.31 / 38.98 \\
XLM-R         & 48.12 / 47.99 & 31.81 / 31.95 \\
mBERT         & 45.74 / 48.49 & 31.07 / 33.02 \\
\bottomrule
\end{tabular}
\end{table}
Overall, AMP effectively maintains task efficacy across sentence-level, token-level, and span-level NLP tasks, while enabling substantial computational savings, which are analyzed further in the training efficiency section.
\subsection{Training Efficiency}
\begin{figure*}[h]
    \centering
    \includegraphics[width=0.99\linewidth]{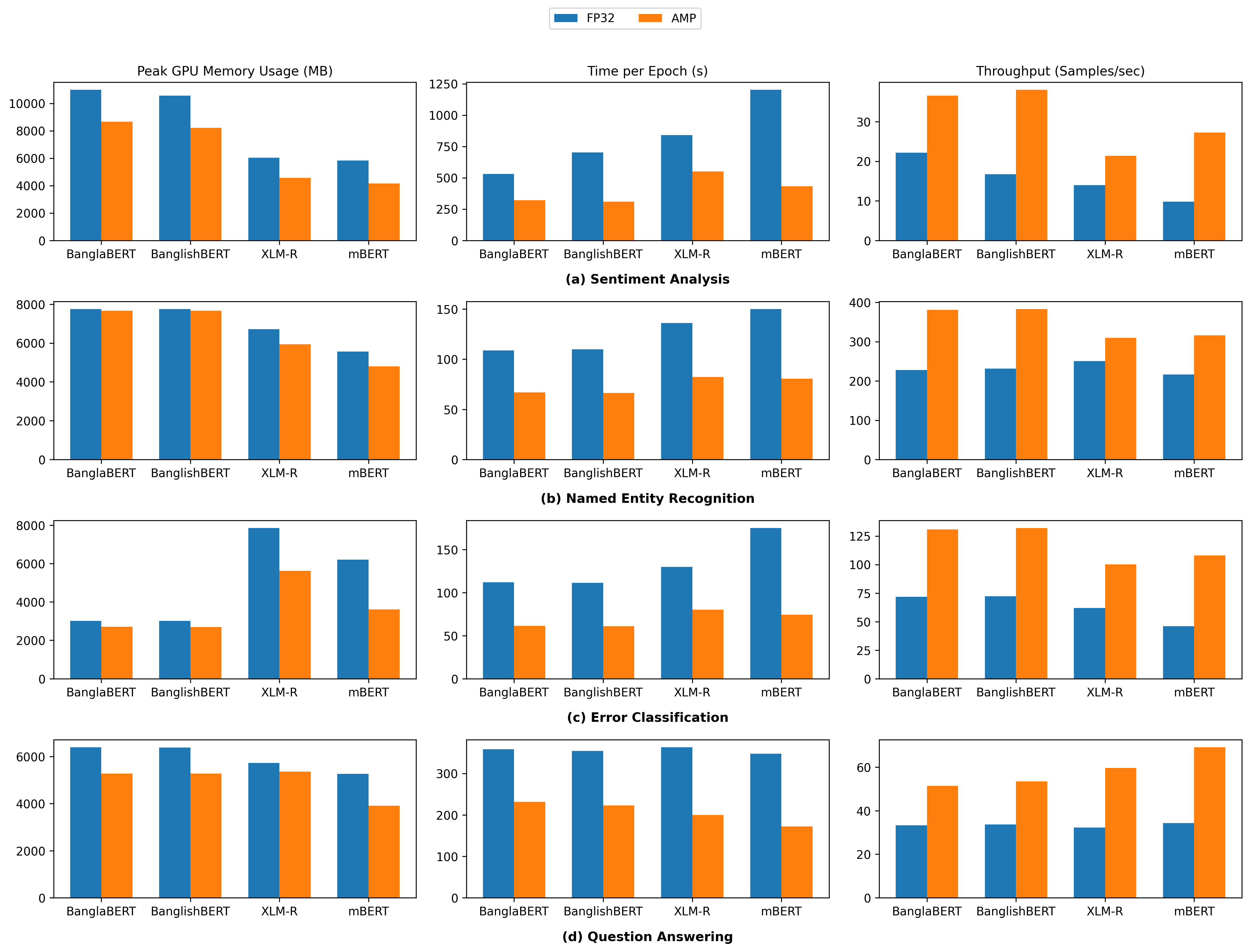}
    \caption{Comparison of FP32 and AMP training efficiency across four Bangla NLP tasks. }
    \label{fig:metrics}
\end{figure*}

We evaluate the computational efficiency gains of Automatic Mixed Precision (AMP) training (FP16/FP32) across four Bangla NLP tasks using pretrained transformer models. Three metrics are considered: (i) peak GPU memory consumption, (ii) throughput measured in samples per second, and (iii) training time per epoch. All experiments are conducted on identical hardware and software to ensure a fair comparison between FP32 and AMP. Results are summarized in Figure~\ref{fig:metrics}.

\textbf{Memory Usage.} Automatic Mixed Precision (AMP) consistently reduces GPU memory consumption across all models and tasks. The task-wise memory reductions range as follows: Sentiment Analysis: 1.2–21.2\%, NER: 1.1–22.2\%, Error Classification: 6.5–28.4\%, QA: 13.9–41.9\%. The largest savings are observed in larger multilingual models, such as XLM-R and mBERT, whereas smaller models, including BanglishBERT, show comparatively modest reductions. These reductions enable larger batch sizes and make training feasible in resource-constrained environments, highlighting AMP’s practical utility for Bangla NLP research.

\textbf{Throughput.} AMP significantly improves training throughput across all tasks and models. The observed task-wise increases in samples per second are as follows: Sentiment Analysis: 54–82\%, NER: 59–127\%, Error Classification: 23–85\%, QA: 46–179\%. The largest gains are observed for QA and NER with larger multilingual models such as XLM-R and mBERT, while smaller monolingual models (BanglaBERT, BanglishBERT) show moderate improvements. For example, Sentiment Analysis throughput increased from 22.2 to 36.6 samples/sec for BanglaBERT (65\%) and from 13.99 to 21.38 samples/sec for XLM-R (82\%). These results demonstrate that AMP not only reduces memory usage but also maximizes hardware utilization, enabling faster model training across diverse Bangla NLP tasks.

\textbf{Training Time.} AMP consistently reduces the average time per epoch across all models and tasks. Task-wise reductions range from 34–64\%, with the largest gains observed in QA and NER. For instance, Sentiment Analysis epoch time decreased from 530.7s to 321.5s for BanglaBERT (39.4\%) and from 840.0s to 549.9s for XLM-R (45.2\%). Similarly, QA training time dropped by 46–64\% depending on the model. Smaller monolingual models like BanglishBERT benefit moderately, while larger multilingual models such as XLM-R and mBERT achieve the highest speedups. These reductions, combined with memory savings and increased throughput, demonstrate AMP’s effectiveness in accelerating Bangla NLP training without compromising model performance.

Overall, AMP delivers substantial efficiency with lower memory consumption, higher throughput, and reduced training time, while preserving model performance. These results highlight the practicality of AMP for Bangla NLP research in hardware-limited settings.

\subsection{Impact of Batch Size on Training Performance}

\begin{figure*}[!t]
    \centering
    \includegraphics[width=\textwidth]{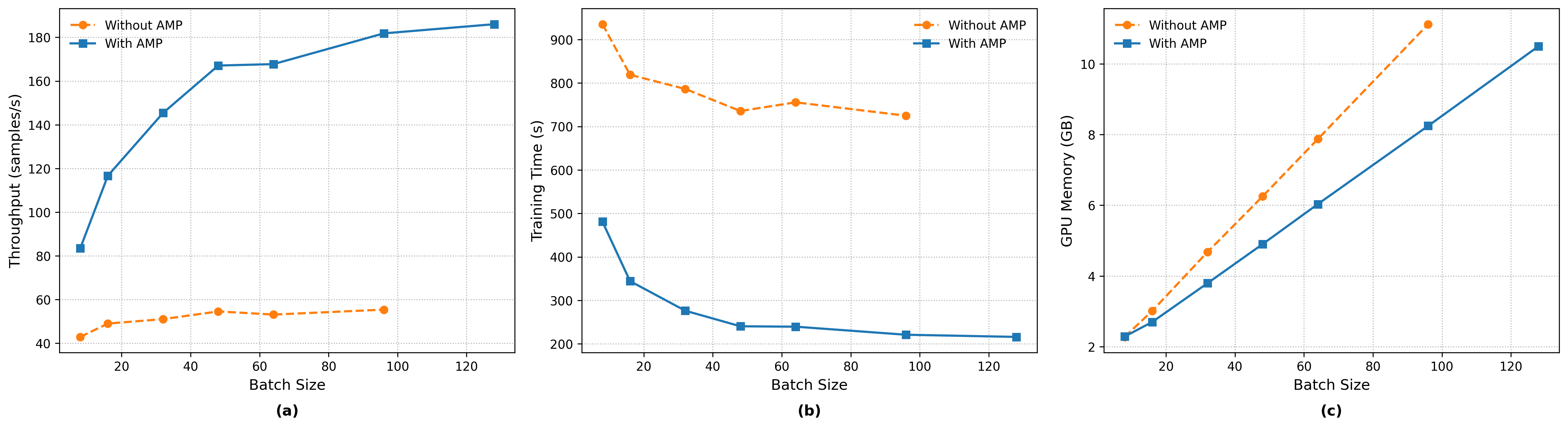}
    \caption{Comparison of AMP and full-precision training across batch sizes. 
    (a) Throughput, (b) Training time, (c) GPU memory usage. AMP substantially improves 
    throughput and reduces training time without significant memory overhead.}
    \label{fig:batch_size}
\end{figure*}

To further explore practical strategies for improving training efficiency, we analyze how varying batch sizes influence performance and resource utilization under FP32 and AMP regimes. We investigate the effect of batch size on training efficiency for the Error Classification task using the BanglaBERT model. Experiments were conducted under both FP32 and Automatic Mixed Precision (AMP), measuring per-epoch training time, throughput, and peak GPU memory across batch sizes of 8, 16, 32, 48, 64, 96, and 128. Figure~\ref{fig:batch_size} shows that AMP enables larger batch sizes without exceeding GPU memory limits. FP32 training failed at batch size 128 due to out-of-memory errors, whereas AMP successfully processed this batch with peak memory usage of 10.5 GB. Across the batch size range, AMP consistently improved efficiency: throughput increased by 95–228\%, per-epoch training time decreased by 49–70\%, and peak memory usage was reduced by up to 26\%.

These findings highlight AMP’s ability to accelerate training while enabling larger batch sizes, improving GPU utilization, and maintaining model performance in resource-constrained environments.

\subsection{Discussion}
Our results demonstrate that Automatic Mixed Precision (AMP) substantially improves computational efficiency for Bangla NLP tasks without compromising performance. Peak GPU memory usage decreased by up to 42\%, training throughput increased by as much as 179\%, and per-epoch training time dropped by up to 64\%. Task metrics, including Accuracy, F1, and Exact Match, remain largely unchanged, with differences within 1–2 percentage points. AMP also enables larger batch sizes, improving GPU utilization and convergence. These findings establish AMP as a practical, efficient approach for training transformer-based Bangla NLP models in resource-constrained environments.

\section{Conclusion}
This research systematically compared full-precision (FP32) and automatic mixed precision (AMP) training across multiple Bangla NLP tasks, including sentiment analysis, named entity recognition, error classification, and question answering. Experimental results demonstrate that AMP substantially reduces peak GPU memory consumption, decreases per-epoch training time, and increases throughput, all without compromising task performance. These findings are particularly valuable for Bangla, where computational resources are often limited, enabling more efficient model development and broader experimentation. Future work includes scaling AMP to larger and multilingual pretrained models, integrating complementary techniques such as gradient checkpointing and quantization, evaluating performance on domain-specific, dialectal, and code-mixed corpora, and exploring energy-efficient training and real-time inference on resource-constrained hardware to advance sustainable and practical Bangla NLP applications.

\bibliographystyle{IEEEtran}
\bibliography{ref}
\end{document}